\documentclass{article}

 \usepackage[preprint]{neurips_2024}

\usepackage[utf8]{inputenc} %
\usepackage[T1]{fontenc}    %

\usepackage{url}            %
\usepackage{booktabs}       %
\usepackage{amsfonts}       %
\usepackage{nicefrac}       %
\usepackage{microtype}      %
\usepackage{xcolor}         %

\definecolor{Klein_Blue}{rgb}{0.0, 0.129, 0.6}
\usepackage{hyperref}       %
\hypersetup{
    colorlinks=true,
    linkcolor=magenta,
    urlcolor=magenta,
    citecolor=Klein_Blue
    }

\usepackage{comment}
\usepackage{array}
\usepackage{graphicx}
\usepackage{amsmath} 
\usepackage{enumitem}
\usepackage{wrapfig}
\usepackage{arydshln} 
\usepackage{xcolor}
\usepackage[normalem]{ulem} %

\definecolor{mycolor1}{HTML}{1F77B4} %
\definecolor{mycolor2}{HTML}{FF7F0E} %
\definecolor{mycolor3}{HTML}{2CA02C} %

\usepackage{xspace}
\newcommand{\taskname}{T2MVid\xspace}

\title{\textcolor{mycolor1}{Vi}\textcolor{mycolor2}{vid}-\textcolor{mycolor3}{ZOO}:  Multi-\textcolor{mycolor1}{Vi}ew \textcolor{mycolor2}{Vid}eo Generation with Diffusion Model}

\author{%
 Bing Li$^*$, Cheng Zheng$^*$, Wenxuan Zhu$^*$, Jinjie Mai, Biao Zhang,\\
\textbf{Peter Wonka,}
\textbf{Bernard Ghanem} \\
King Abdullah University of Science and Technology \\
 \texttt{\{bing.li,bernard.ghanem\}@kaust.edu.sa} \\
}

\newcommand{\ie}{\textit{i}.\textit{e}., }
\newcommand{\eg}{\textit{e}.\textit{g}., }
\newcommand{\etal}{\textit{et al.}}
\newcommand{\expimgw}{0.105\textwidth}

\begin{document}

\maketitle

\begin{abstract}

While diffusion models have shown impressive performance in 2D image/video generation, diffusion-based Text-to-Multi-view-Video (\taskname) generation remains underexplored.
The new challenges posed by \taskname generation lie in the lack of massive captioned multi-view videos and the complexity of modeling such multi-dimensional distribution.
To this end, we propose a novel diffusion-based pipeline that generates high-quality multi-view videos centered around a dynamic 3D object from text. 
Specifically, we factor the \taskname problem into viewpoint-space and time components. 
Such factorization allows us to combine and reuse layers of advanced pre-trained multi-view image and 2D video diffusion models to ensure multi-view consistency as well as temporal coherence for the generated multi-view videos, largely reducing the training cost. 
 We further introduce alignment modules to align the latent spaces of layers from the pre-trained multi-view and the 2D video diffusion models, addressing the reused layers' incompatibility that arises from the domain gap between 2D and multi-view data. 
To facilitate this research line,  we further contribute a captioned multi-view video dataset.
Experimental results demonstrate that our method generates high-quality multi-view videos, exhibiting vivid motions, temporal coherence, and multi-view consistency, given a variety of text prompts. Project page is at \href{https://hi-zhengcheng.github.io/vividzoo}{\texttt{\textcolor{purple}{\uline{vividzoo}}}}.

\end{abstract}

\section{Introduction}

\begin{figure*}[t]
\centering
\tabcolsep=0.05cm
\begin{tabular}{ccccccc}
\centering
\rotatebox{90}{\xspace\xspace View 0} 
& {\includegraphics[bb=0 0 256 256,width=0.13\textwidth]{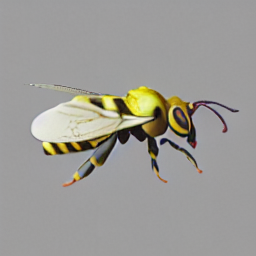}}
& {\includegraphics[bb=0 0 256 256,width=0.13\textwidth]{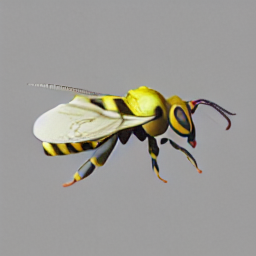}}
& {\includegraphics[bb=0 0 256 256,width=0.13\textwidth]{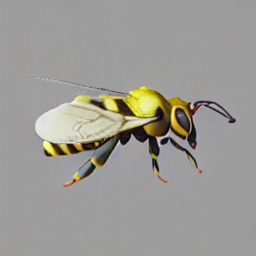}}
& {\includegraphics[bb=0 0 256 256,width=0.13\textwidth]{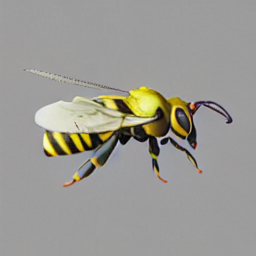}}
& {\includegraphics[bb=0 0 256 256,width=0.13\textwidth]{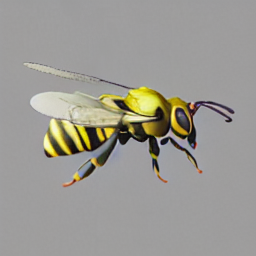}}
& {\includegraphics[bb=0 0 256 256,width=0.13\textwidth]{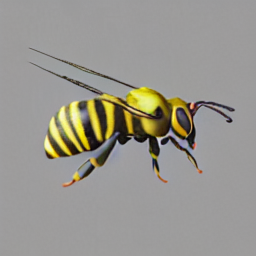}}
\\
\rotatebox{90}{\xspace\xspace View 1} 
& {\includegraphics[bb=0 0 256 256,width=0.13\textwidth]{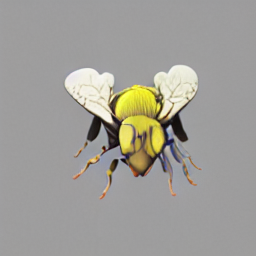}}
& {\includegraphics[bb=0 0 256 256,width=0.13\textwidth]{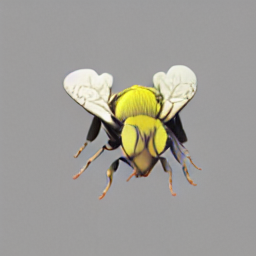}}
& {\includegraphics[bb=0 0 256 256,width=0.13\textwidth]{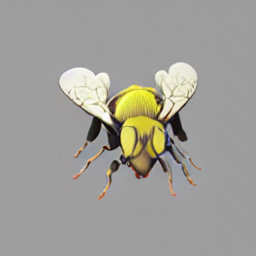}}
& {\includegraphics[bb=0 0 256 256,width=0.13\textwidth]{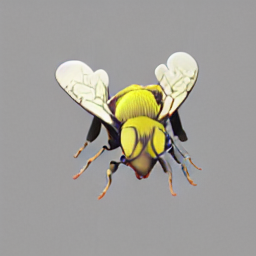}}
& {\includegraphics[bb=0 0 256 256,width=0.13\textwidth]{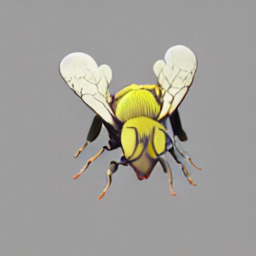}}
& {\includegraphics[bb=0 0 256 256,width=0.13\textwidth]{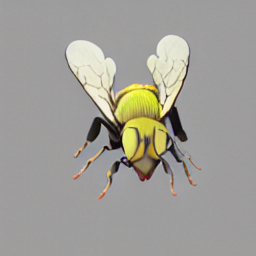}}
\\
\rotatebox{90}{\xspace\xspace View 2} 
& {\includegraphics[bb=0 0 256 256,width=0.13\textwidth]{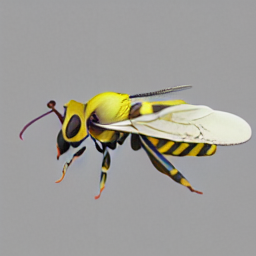}}
& {\includegraphics[bb=0 0 256 256,width=0.13\textwidth]{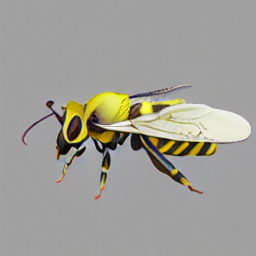}}
& {\includegraphics[bb=0 0 256 256,width=0.13\textwidth]{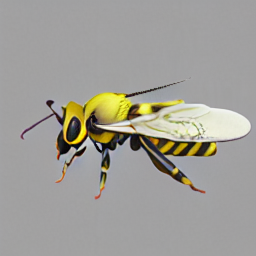}}
& {\includegraphics[bb=0 0 256 256,width=0.13\textwidth]{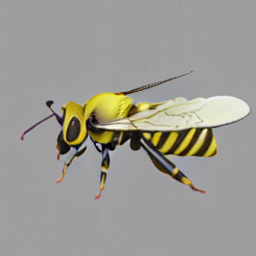}}
& {\includegraphics[bb=0 0 256 256,width=0.13\textwidth]{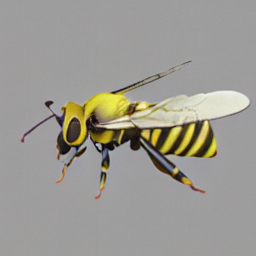}}
& {\includegraphics[bb=0 0 256 256,width=0.13\textwidth]{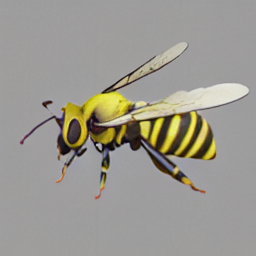}}
\\
\rotatebox{90}{\xspace\xspace View 3} 
& {\includegraphics[bb=0 0 256 256,width=0.13\textwidth]{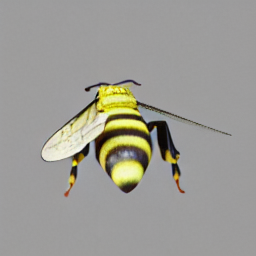}}
& {\includegraphics[bb=0 0 256 256,width=0.13\textwidth]{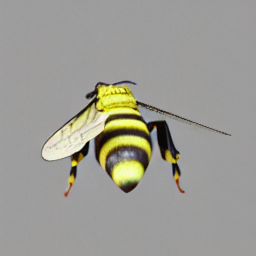}}
& {\includegraphics[bb=0 0 256 256,width=0.13\textwidth]{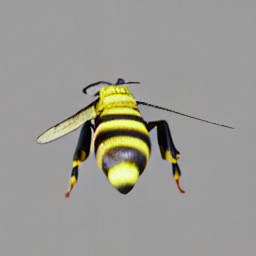}}
& {\includegraphics[bb=0 0 256 256,width=0.13\textwidth]{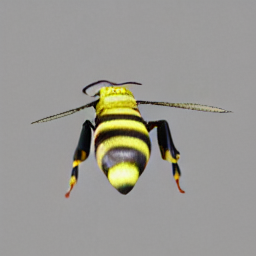}}
& {\includegraphics[bb=0 0 256 256,width=0.13\textwidth]{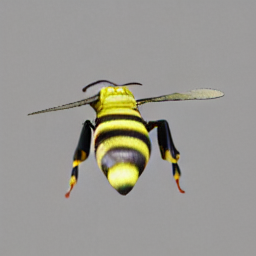}}
& {\includegraphics[bb=0 0 256 256,width=0.13\textwidth]{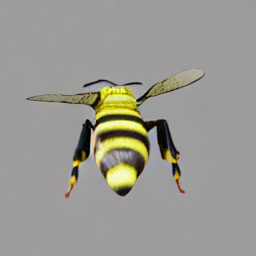}}

\end{tabular}
\\

{\textit{Text prompt: a yellow and black striped wasp bee, 3d asset}}
\caption{The proposed Vivid-ZOO generates high-quality multi-view videos of a dynamic 3D object from text. Each row illustrates six frames drawn from a generated video for one viewpoint.}
\vspace{-15pt}
\label{fig:bee}
\end{figure*}

Multi-view videos capture a scene/object from multiple cameras with different poses simultaneously, which are critical for numerous downstream applications~\cite{4d_vis,mvid_system,mvid_recon} such as AR/VR, 3D/4D modeling, media production, and interactive entertainment.
More importantly, the availability of such data holds substantial promise for facilitating progress in research areas such as 4D reconstruction~\cite{neural_3d_video,lin2023im4d}, 4D generation~\cite{4dfy, alignyourgaussians}, and long video generation~\cite{sora,long_form_video} with 3D consistency. 
However, collecting multi-view videos often requires sophisticated setups~\cite{mamal_pig} to synchronize and calibrate multiple cameras, resulting in a significant absence of datasets and generative techniques for multi-view videos.

In the meantime, diffusion models have shown great success in 2D image/video generation. For example, 2D video diffusion models \cite{svd, animatediff,vdm,makeavideo}  generate high-quality 2D videos by extending image diffusion models \cite{ldm, sdv2}. 
Differently, multi-view image diffusion models \cite{MVDream,spad,wonder3d,imagedream} are proposed to generate multi-view images of 3D objects, which have demonstrated significant impact in 3D object generation \cite{li2024mvcontrol}, 3D reconstruction \cite{Magic123}, and related fields. 
However, to the best of our knowledge, no other works have explored Text-to-Multi-view-Video (\taskname) diffusion models.
Motivated by recent 2D video and multi-view image diffusion models, we aim to propose a diffusion-based method that generates multi-view videos of dynamic objects from text (see Fig. \ref{fig:bee}). 

Compared to 2D video generation, \taskname generation poses two new challenges. 
First, modeling multi-view videos is complex due to their four-dimensional nature, which involves different viewpoints as well as the dimensions of time and space (2D).
Consequently, it is nontrivial for diffusion models to model such intricate data from scratch without extensive captioned multi-view video datasets.
Second, there are no publicly available large-scale datasets of captioned multi-view videos, but it has been shown that billions of text and 2D image pairs are essential for powerful image diffusion models \cite{ldm, sd15, sdv2}. For example, Stable Diffusion \cite{sd15} is trained on the massive LAION-5B dataset \cite{laion5b}.
Unlike downloading 2D images available on the Internet, collecting a large quantity of multi-view videos is labor-intensive and time-consuming. This challenge is further compounded when high-quality captioned videos are needed, hindering the extension of diffusion models to \taskname generation.

In this paper, instead of the labor-intensive task of collecting a large amount of captioned multi-view video data, 
we focus on the problem of enabling diffusion models to generate multi-view videos from text using only a comparable small dataset of captioned multi-view videos. This problem has not been taken into account by existing diffusion-based methods (\eg \cite{svd}\cite{animatediff}). 
However, studies have revealed that naively fine-tuning a large pre-trained model on limited data can result in overfitting \cite{lora, dreambooth, controlnet}.  
Our intuition is that we can factor the multi-view video generation problem into viewpoint-space and time components. 
The viewpoint-space component ensures that the generated multi-view videos are geometrically consistent and aligned with the input text, and the temporal component ensures temporal coherence.
With such factorization, a straightforward approach is to leverage large-scale multi-view image datasets (\eg \cite{zero123} \cite{richdreamer}) and 2D video datasets (\eg Web10M \cite{Web10M}) to pre-train the viewpoint-space component and temporal component, respectively.  
However, while this approach can largely reduce the reliance on extensive captioned multi-view videos, it remains costly in terms of training resources.

Instead,  we explore a new question: \textit{ can we jointly combine and reuse the layers of pre-trained 2D video and multi-view image diffusion models to establish a \taskname diffusion model}?  The large-scale pre-trained multi-view image diffusion models (\eg MVdream \cite{MVDream}) have learned how to model multi-view images, and the 2D temporal layers of powerful pre-trained video diffusion models (\eg AnimateDiff \cite{animatediff}) learned rich motion knowledge. However,  new challenges are posed. We observe that naively combining the layers from these two kinds of diffusion models leads to poor generation results. More specifically,  the training data of multi-view image diffusion models are mainly rendered from synthetic 3D objects (\eg Objaverse \cite{objaverse,objaverseXL} ), while 2D video diffusion models are mainly trained on real-world 2D videos, posing a large domain gap issue.

To bridge this gap, we propose a novel diffusion-based pipeline, namely, \textcolor{mycolor1}{Vi}\textcolor{mycolor2}{vid}-\textcolor{mycolor3}{ZOO}, for \taskname generation.
The proposed pipeline effectively connects the pre-trained multi-view image diffusion model \cite{MVDream} and 2D temporal layers\footnote{For clarification, we add ``2D" when referring to the layers of the 2D video diffusion models, while we add ``3D" when referring to the multi-view image diffusion model.
} of the pre-trained video model by introducing two kinds of layers, named 3D-2D alignment layers and 2D-3D alignment layers, respectively.  
The 3D-2D alignment layers are designed to align features to the latent space of the pre-trained 2D temporal layers, and the introduced 2D-3D alignment layers project the features back.
Furthermore, we construct a comparable small dataset consisting of 14,271 captioned multi-view videos to facilitate this and future research line. 
Although our dataset is much smaller compared to the billion-scale 2D image dataset (LAION \cite{laion5b}) and the million-scale 2D video dataset (e.g., WebVid10M \cite{Web10M}), our pipeline allows us to effectively train a large-scale \taskname diffusion model using such limited data.
Extensive experimental results demonstrate that our method effectively generates high-quality multi-view videos given various text prompts.

We summarize our contributions as follows:
\begin{itemize}[leftmargin=*]
\item We present a novel diffusion-based pipeline that generates high-quality multi-view videos from text prompts. {This is the first study on \taskname diffusion models.}
 \item We show how to combine and reuse the layers of the pre-trained 2D video and multi-view image diffusion models for a \taskname diffusion model.
 The introduced 3D-2D alignment and 2D-3D alignment are simple yet effective, enabling our method to utilize layers from the two diffusion models across different domains, ensuring both temporal coherence and multi-view consistency.

 \item We contribute a multi-view video dataset that provides multi-view videos, text descriptions, and corresponding camera poses, which helps to advance the field of \taskname generation. 

\end{itemize}

\section{Related work}

\textbf{2D video diffusion model.} Many previous approaches have explored autoregressive transformers (\eg \cite{cogview2,cogvideo,videogpt}) or GANs (\eg \cite{brooks2022generating,luc2020transformation,saito2020train})  for video generation.  Recently, more and more efforts have been devoted to diffusion-based video generation \cite{tokenflow2023, imagen, qi2023fatezero, wang2024360dvd, artv, tunevideo, yang2023rerender, magicvideo, kim2024fifodiffusion}, inspired by the impressive results of image diffusion models \cite{pixartdelta, pixartalpha, ldm, sd15, sdv2, controlnet}.

The amount of available captioned 2D video data is significantly less than the vast number of 2D image-text pairs available on the Internet.
Most methods \cite{blattmann2023align, emu, animatediff, genlgvideo, animatelcm, modelscope}  extend pre-trained 2D image diffusion models to video generation to address the challenge of limited training data.
Some methods employ pre-trained 2D image diffusion models (\eg \cite{sd15}) to generate 2D video from texts in a zero-shot manner \cite{text2videozero}\cite{controlvideo} or using few-shot tuning strategies \cite{wu2024lamp}.
These methods avoid the requirement of large-scale training data.
Differently, another research line is to augment pre-trained 2D image diffusion models with various temporal modules or trainable parameters, showing impressive temporal coherence performance. 
For example, 
Ho et al. \cite{vdm} extend the standard image diffusion architecture by inserting a temporal attention block.
Animatediff\cite{animatediff} and AYL \cite{blattmann2023align} freeze 2D image diffusion model and solely train additional motion modules on large-scale datasets of captioned 2D videos such as WebVid10M \cite{Web10M}.
In addition,  image-to-2D-video generation methods  \cite{svd,ren2024consisti2v,dynamicrafter,vgenxl} are proposed based on diffusion models to generate a monocular video from an image.
Methods \cite{videocomposer} focus on controllable video generation through different conditions such as pose and depth. MotionCtrl~\cite{motionctrl} and Direct-a-video~\cite{direct-a-video} can generate videos conditioned by the camera and object motion. CameraCtrl~\cite{cameractrl} can also control the trajectory of a moving camera for generated videos.
However, these  text-to-2D-video diffusion models are designed for monocular video generation, which does not explicitly consider the spatial 3D consistency of multi-view videos.

\textbf{Multi-view image diffusion model.} Recent works have extended 2D image diffusion models for multi-view image generation.
Zero123 \cite{zero123} and Zero123++ \cite{zero123plus} propose to fine-tune an image-conditioned diffusion model so as to generate a novel view from a single image. 
Inspired by this, many novel view synthesis methods \cite{cat3d, jiang2024efficientdim, syncdreamer, wonder3d, im3d, imagedream, wang2024crm, harmonyview, consistnet, consistent1to3, free3D} are proposed based on image diffusion models.
For example, IM3D \cite{im3d} and Free3D \cite{free3D} generate multiple novel views simultaneously to improve spatial 3D consistency among different views.
Differently, a few methods \cite{v3d,vfusion3d,sv3d} adapt pre-trained video diffusion models (\eg \cite{svd}) to generate multi-view images from a single image.
MVDream \cite{MVDream} presents a text-to-multi-view-image diffusion model to generate four views of an object each time given a text, while SPAD \cite{spad} generates geometrically consistent images for more views.  Richdreamer \cite{richdreamer} trains a diffusion model to generate depth, normal, and albedo.

\textbf{4D generation using diffusion models.} 
Many approaches \cite{magic3d, zero123, dreamfusion, DreamBooth3D, zero123plus, MVDream, dreamgaussian, prolificdreamer,tc4d} have exploited pre-trained diffusion models to train 3D representations for 3D object generation via score distillation sampling \cite{dreamfusion}.
Recently, a few methods \cite{4dfy, alignyourgaussians, dreamgaussian4d, makeav3d, 4dgen} leverage pre-trained diffusion models to train 4D representations for dynamic object generation.   For example, Ling \etal \cite{alignyourgaussians} represent a 4D object as Gaussian spatting \cite{gaussiansplatting}, while Bahmani \etal \cite{4dfy} adopt a NeRF-based representation \cite{nerf,instantngp,suds}. Then, pre-trained 2D image, 2D video, and multi-view image diffusion models are employed to jointly train the 4D representations. In addition, diffusion models are used to generate 4D objects from monocular videos \cite{dreamscene4d, jiang2024consistentd}.
Different from all these methods that directly pre-trained diffusion models, our approach focuses on presenting a \taskname diffusion model. 
LMM~\cite{large_motion_model} generates 3D motion for given 3D human models. DragAPart~\cite{drag_a_part} can generate part-level motion for articulated objects.

\section{Multi-view video diffusion model}
\label{sec:method}

\textbf{Problem definition.}
Our goal for \taskname generation is to generate a set of multi-view videos centered around a dynamic object from a text prompt. Motivated by the success of diffusion models in 2D video/image generation, we aim to design a \taskname diffusion model. However,  \taskname generation is challenging due to the complexity of modeling multi-view videos and the difficulty of collecting massive captioned multi-view videos for training.

We address the above challenges by exploring two questions. (1) { Can we design a diffusion model that effectively learns \taskname generation, yet only needs a comparable small dataset of multi-view video data?} 
(2) Can we jointly leverage, combine, and reuse the layers of pre-trained 2D video and multi-view image diffusion models to establish a \taskname diffusion model? Addressing these questions can reduce the reliance on large-scale training data and decrease training costs. However, this question remains unexplored for diffusion-based \taskname.

\begin{figure*}[t]
    \centering
    \includegraphics[width=1.0\textwidth]{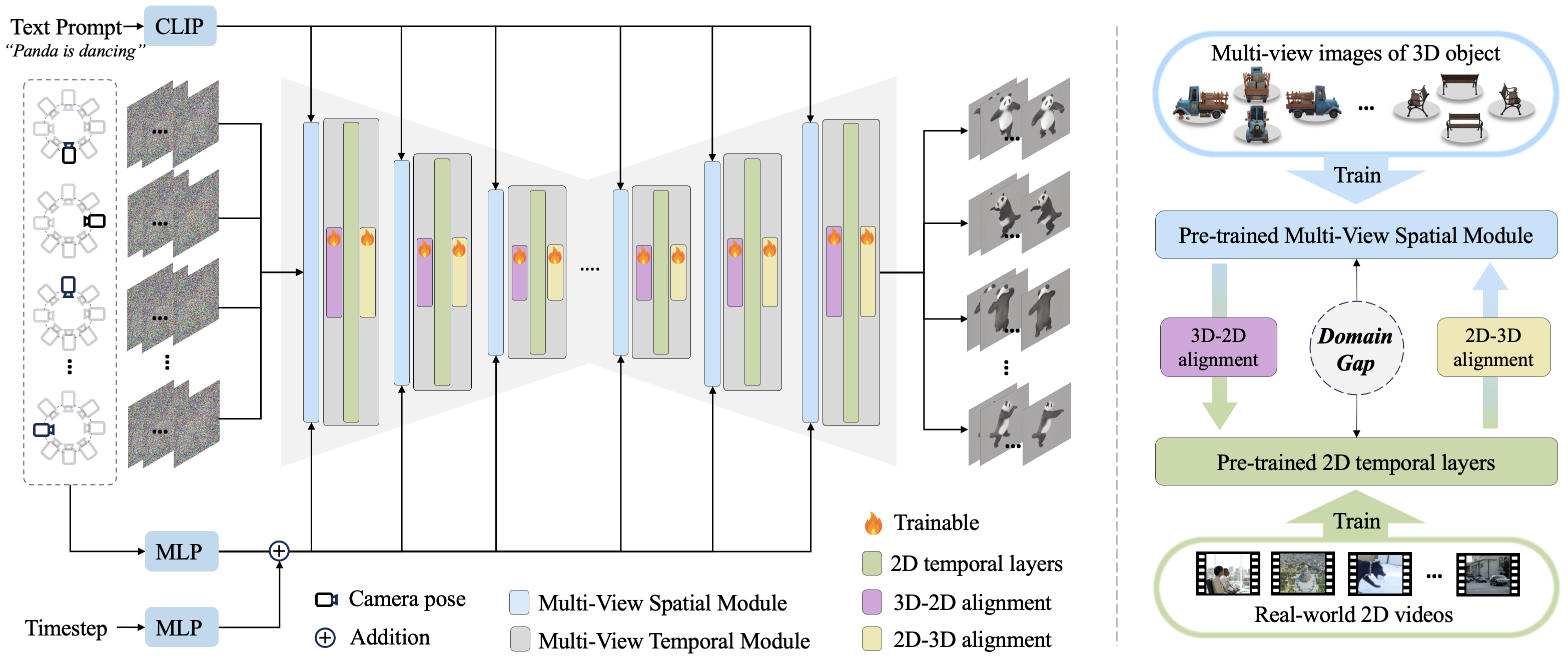}

\caption{Overview of the proposed Vivid-ZOO. {\textbf{Left}}: Given a text prompt, our diffusion model generates multi-view videos. Instead of training from scratch, the multi-view spatial module reuses the pre-trained multi-view image diffusion model, and the multi-view temporal module leverages the 2D temporal layers of the pre-trained 2D video diffusion model to enforce temporal coherence. \textbf{Right}: Jointly reusing the pre-trained multi-view image diffusion model and temporal 2D layers poses new challenges due to the large gap between their training data (multi-view images of synthetic 3D objects versus real-world 2D videos). We introduce 3D-2D alignment and 2D-3D alignment to address the domain gap issue.}

    \label{fig:network}
    \vspace{-16pt}
\end{figure*}

\textbf{Overview.} We address the above questions by factoring the \taskname generation problem over viewpoint-space and time. With the factorization, we propose a diffusion-based pipeline for \taskname generation  (see Fig \ref{fig:network}),  including the multi-view spatial modules and multi-view temporal modules. Sec \ref{sec:spatial} describes how we adapt a pre-trained multi-view image diffusion model as the multi-view spatial modules. Multi-view temporal modules effectively leverage temporal layers of the pre-trained 2D video diffusion model with the newly introduced 3D-2D alignment layers and 2D-3D alignment layers (Sec \ref{set:temporal}). Finally, we describe training objectives in Sec \ref{sec:training_objective} and the dataset construction to support our pipeline for \taskname generation in Sec \ref{sec:dataset}.

\label{sec:framework}
\subsection{Multi-view spatial module}
\label{sec:spatial}

Our multi-view spatial modules ensure that the generated multi-view videos are geometrically consistent and aligned with the input text.
Recent multi-view image diffusion models \cite{spad, MVDream} generate high-quality multi-view images by fine-tuning Stable Diffusion and modifying its self-attention layers.
We adopt the architecture of Stable Diffusion for our multi-view spatial modules. Furthermore, we leverage a pre-trained multi-view image diffusion model based on Stable Diffusion by reusing its pre-trained weights in our spatial modules, which avoids training from scratch and reduces the training cost. 
However, the self-attention layers of Stable Diffusion are not designed for multi-view videos. We adapt these layers for multi-view self-attention as below.

\textbf{Multi-view self-attention.}  We inflate self-attention layers to capture geometric consistency among generated multi-view videos. Let $\mathbf{F} \in \mathbb{R}^{b\times K\times N\times d\times h\times w}$ denote the 6D feature tensor of multi-view videos in the diffusion model, where $b$, $K$, $N$, $d$ and $h\times w$ are batch size, view number, frame number, feature channel and spatial dimension, respectively. Inspired by  \cite{spad, MVDream},  we reshape $\mathbf{F}$ into a shape of ${(b\times N)\times d  \times (K\times h\times w)}$, leading to a batch of feature maps $\mathbf{\Tilde{F}}^{n}$ of 2D images, where $(b\times N)$ is the batch size, $\mathbf{\Tilde{F}}^{n}$ denotes a feature map representing  all views at frame index $n$,  and $(K\times h \times w)$ is the spatial size. We then feed the reshaped feature maps  $\mathbf{\Tilde{F}}^{n}$ into self-attention layers. Since $\mathbf{\Tilde{F}}^{n}$ consists of all views at frame index $n$, the self-attention layers learn to capture geometrical consistency among different views. We also inflate other layers of stable diffusion (see Appendix~\ref{apx:implementation_details}) so that we can reuse their pre-trained weight.

\textbf{Camera pose embedding.} 
Our diffusion model is controllable by camera poses, achieved by incorporating a camera pose sequence as input.
These poses are embedded by MLP layers and then added to the timestep embedding, following MVdream \cite{MVDream}. Here, our multi-view spatial module reuses the pre-trained multi-view image diffusion model MVDream \cite{MVDream}.

\subsection{Multi-view temporal module}
\label{set:temporal}
Besides spatial 3D consistency, it is crucial for \taskname diffusion models to maintain the temporal coherence of generated multi-view videos simultaneously.
Improper temporal constraints would break the synchronization among different views and introduce geometric inconsistency. Moreover, training a complex temporal module from scratch typically requires a large amount of training data.

\begin{wrapfigure}[21]{r}{0.38\textwidth}
\vspace{-58pt}
\centering
\begin{minipage}[t]{0.76\textwidth}
    \includegraphics[bb=0 0 1462 1382, width=0.67\textwidth]{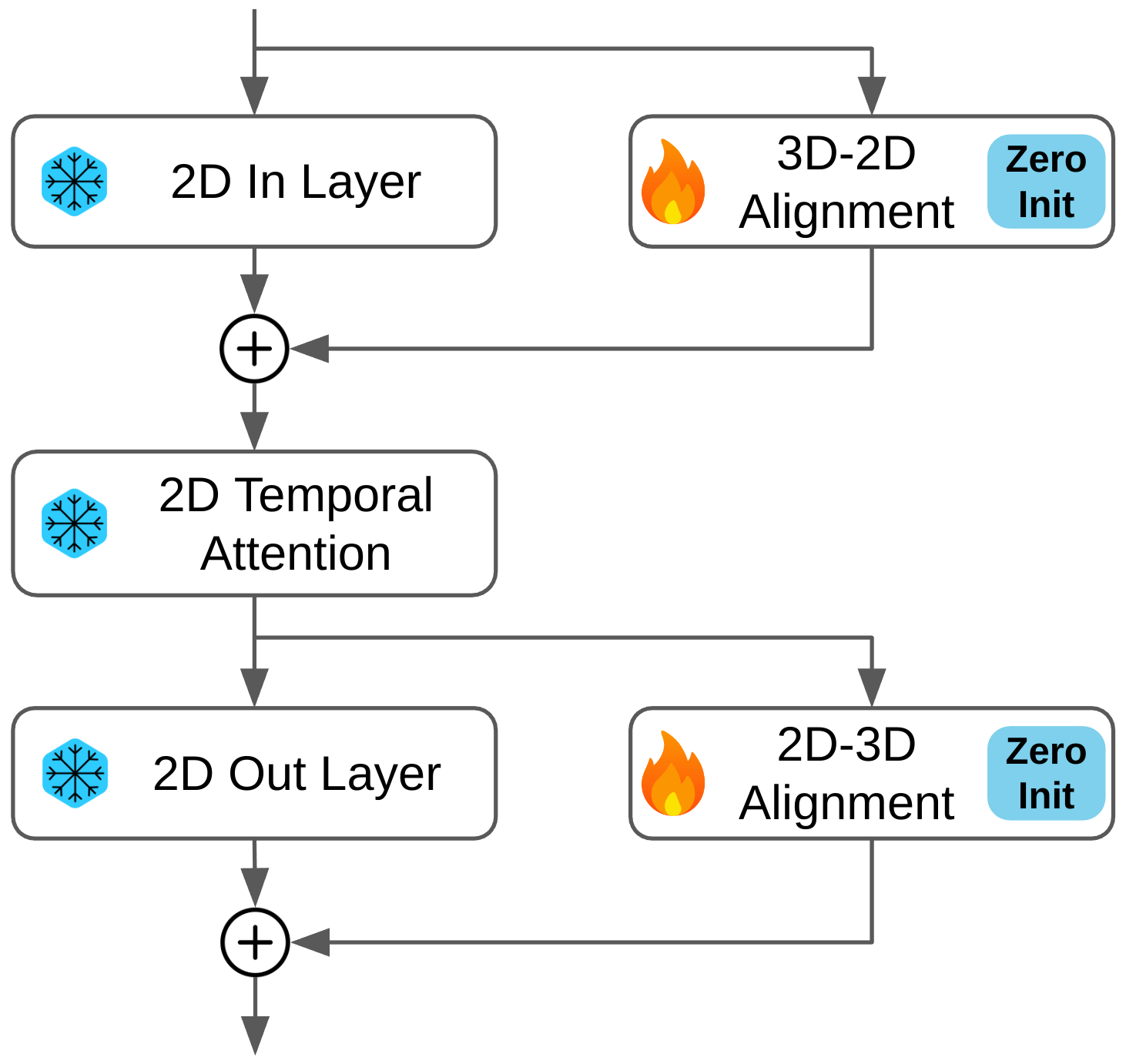}
    \end{minipage}
\caption{Our multi-view temporal module, where 3D-2D alignment layers are trained to align features to the latent space of the 2D temporal attention layers, and the 2D-3D alignment layers project them back. }
\label{fig:me_temporal}	
\end{wrapfigure}

Instead, we propose to leverage the 2D temporal layers of large pre-trained 2D video diffusion models (\eg \cite{animatediff}) to ensure temporal coherence for \taskname generation. These 2D temporal layers have learned rich motion priors, as they have been trained on millions of 2D videos (\eg \cite{Web10M}). Here,  we employ the 2D temporal layers of AnimateDiff \cite{animatediff} due to its impressive performance in generating temporal coherent 2D videos.

However, we observed that naively combining the pre-trained 2D temporal layers with the multi-view spatial module leads to poor results. The incompatibility is due to the fact that the pre-trained 2D temporal layers and the multi-view spatial modules are trained on data from different domains (\ie real 2D and synthetic multi-view data) that have a large domain gap. 
To address the domain gap issue, one approach is to fine-tune all 2D temporal layers of a pre-trained 2D video diffusion model on multi-view video data. However, such an approach not only needs to train many parameters but can also harm the learned motion knowledge  \cite{lora} if a small training dataset is given.  We present a multi-view temporal module (see Fig. \ref{fig:me_temporal}) that reuses and freezes all 2D temporal layers to maintain the learned motion knowledge and introduce the 3D-2D alignment layer and the 2D-3D alignment layer.

\textbf{3D-2D alignment.} We introduce the 3D-2D alignment layers to effectively combine the pre-trained 2D temporal layers with the multi-view spatial module. Recently, a few methods \cite{animatediff,svd} add motion LoRA to 2D temporal attention for personalized/customized video generation tasks.
However, our aim is different, \ie we expect to preserve the learned motion knowledge of 2D temporal layers, such that our multi-view temporal module can leverage the knowledge for ensuring temporal coherence. 

Since motion prior knowledge is captured by the pre-trained 2D temporal attention layers, we insert the 3D-2D alignment layers before the 2D temporal attention layers. The 3D-2D alignment layers are learned to align the features into the latent space of the pre-trained 2D temporal layers. Furthermore, inspired by ControlNet \cite{controlnet} and \cite{resnet}, the 3D-2D alignment layers are inserted via residual connections and are zero-initialized, providing an identity mapping at the beginning of training.
The process is described as follows:
\begin{align}
\mathbf{F}^{} =\alpha^{2D}(\mathbf{F}) +\alpha^{3D\rightharpoonup2D}(\mathbf{F}) 
\end{align}
where $\alpha^{3D\rightharpoonup2D}$  is the 3D-2D alignment layer. $\alpha^{2D}$ is the 2D temporal layer followed by the 2D temporal attention layers and we refer to it as \textit{2D in-layer} (see more details in Appendix). 
The 3D-2D alignment layer is plug-and-play and is simply implemented as an MLP.

\textbf{Multi-view temporal coherence.} We reuse and freeze the pre-trained 2D temporal layers in our multi-view temporal module to ensure the temporal coherence of each generated video. 
However, the 2D temporal layer is designed to handle 2D videos. We inflate the 2D temporal layer by reshaping the feature $\mathbf{F}$ to the 2D video dimension via the \textit{rearrange} operation \cite{rearrange}. Then, 2D temporal layers $\gamma (\cdot)$ model temporal coherence across frames by  calculating the attention of points at the same spatial location  in $\mathbf{F}$ across frames for each video: 
\begin{align}
&  \mathbf{F} = \text{rearrange}(\mathbf{F}, ~ b~K~N~h~w~d \rightarrow  (b~K~h~w)~ N~d)\\
&  \mathbf{F} =  \gamma(\mathbf{F}) \\
&  \mathbf{F} = \text{rearrange}(\mathbf{F},   (b~K~h~w)~ N~d \rightarrow ~ b~K~N~h~w~d )
\end{align}

\textbf{2D-3D alignment.}
 We add the 2D-3D alignment layers after 2D temporal attention layers to project the feature back to the feature space of the multi-view spatial modules. %
\begin{align}
&  \mathbf{F}^a = \beta^{2D}(\mathbf{F}) + \beta^{2D\rightharpoonup 3D}(\mathbf{F}) 
\end{align}
where $\beta^{3D\rightharpoonup2D}$  is the 2D-3D alignment layer. $\beta^{2D}$ is the 2D temporal layer following the 2D temporal attention layer. The 2D-3D alignment layers are implemented as an MLP.

\subsection{Training objectives} 
\label{sec:training_objective}

We train our diffusion model to generate multi-view videos. 
Note that we freeze most layers/modules in the diffusion model and only train the 3D-2D and 2D-3D alignment layers during training, which largely reduces the training cost and reliance on large-scale data.
Let $\mathcal{X}$ denote the training dataset, where a training sample $\{\mathbf{x},y, \mathbf{c} \}$ consists of $N$ multi-view videos $\mathbf{x}=\{x\}_1^N$, $N$ corresponding camera poses $\mathbf{c}$, and a text prompt $y$. The training objective $\mathcal{L}$ on $\mathcal{X}$ is defined as follows:
\begin{align}
   & \mathcal{L} = \mathbb{E}_{\mathbf{z}_t^v,y,\epsilon,t} \left[ \left \| \epsilon - \epsilon_{\theta}(\mathbf{z}_t^v, t, \tau_\theta(y), \mathbf{c}) \right\|^2 \right]  
\end{align}
where $\tau_\theta(\cdot)$ is a text encoder that encodes the text into text embedding, $\epsilon_\theta(\cdot)$ is the denoising network. $\mathbf{z}_0^v$ is the latent code of a multi-view video sequence and $\mathbf{z}_t^v$ is its noisy code with added noise $\epsilon$.

\subsection{Multi-view video dataset}
\label{sec:dataset}

Different from 2D images that are available in vast numbers on the Internet, it is much more difficult and expensive to collect a large amount of multi-view videos centered around 3D objects and corresponding text captions.
Recently, multi-view image datasets (\eg \cite{zero123,richdreamer}), rendered from synthetic 3D models, have shown a significant impact on various tasks such as novel view synthesis \cite{zero123, imagedream}, 3D generation (Gaussian Splatting \cite{tang2024lgm}, large reconstruction model \cite{one2345}) and multi-view image generation \cite{MVDream} and associated applications.
Motivated by this, we resort to rendering multi-view videos from synthetic 4D models (animated 3D models).  

We construct a dataset named MV-VideoNet that provides 14,271 triples of a multi-view video sequence, its associated camera pose sequence, and a text description. In particular,  we first select animated objects from Objaverse \cite{objaverse}.  Objaverse is an open-source dataset that provides high-quality 3D objects and animated ones (\ie 4D object). We select 4D objects from the Objaverse dataset and discard those without motions or with imperceptible motions. 
Given each selected 4D object, we render 24-view videos from it, where the azimuth angles of camera poses are uniformly distributed.  To improve the quality of our dataset,  we manually filter multi-view videos with low-quality  \eg distorted shapes or motions, very slow or rapid movement. 
For text descriptions,  we adopt the captioning method Cap3D \cite{c3d2, c3d} to caption a multi-view video sequence. Cap3D leverages BLIP2 \cite{BLIP2} and GPT4-Vision \cite{openai2024gpt4} to fuse information from multi-view images, generating text descriptions.

\section{Experiments}

\textbf{Implementation details.} We reuse the pre-trained MVDream V1.5 in our multi-view spatial module and reuse the pre-trained 2D temporal layers of AnimateDiff V2.0 in our multi-view temporal module.   
We train our model using AdamW \cite{admw} with a learning rate of $10^{
-4}$. During training, we process the training data by randomly sampling 4 views that are orthogonal to each other from a multi-view video sequence, reducing the spatial resolution of videos to $256\times 256$,  and sample video frames with a stride of 3. Following AnimateDiff, we use a linear beta schedule with $\beta_{start}$ = 0.00085 and $\beta_{end}$ = 0.012. (Please refer to the Appendix for more details).

\textbf{Evaluation metrics.}
 Quantitatively evaluating multi-view consistency and temporal coherence remains an open problem for \taskname generation.
We quantitatively evaluate text alignment via CLIP \cite{clip} and temporal coherence via  {Frechet Video Distance (FVD) \cite{FVD}}. Moreover,  we conduct a user study to evaluate the overall performance incorporating text alignment, temporal coherence, and multi-view consistency according to human preference (H. Pref.).

\subsection{Qualitative and quantitative results}
 
To the best of our knowledge, no studies have explored \taskname diffusion models before. We establish a baseline method named {\textbf{MVDream + IP-AnimateDiff}} for comparison.  
{\textbf{MVDream + IP-AnimateDiff}} combines the pre-trained multi-view image diffusion model {\textit{MVDream}} \cite{MVDream} and the 2D video diffusion model {\textit{AnimateDiff}} \cite{animatediff}, since  MVDream generates high-quality multi-view images and AnimateDiff generates temporal coherent 2D videos. Following \cite{pia}, we combine AnimateDiff with IP-adaptor \cite{ipadaptor} to enable AnimateDiff to take an image as input.

\begin{figure*}[t]
\centering
\tabcolsep=0.02cm
\begin{tabular}{cc:cccc|cccc}
&MVdream&\multicolumn{4}{c|}{MVdream + IP-AnimateDiff} & \multicolumn{4}{c}{Ours} \\
\rotatebox{90}{\xspace\xspace View 0} 
& {\includegraphics[bb=0 0 256 256,width=\expimgw]{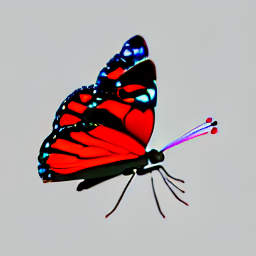}}

& {\includegraphics[bb=0 0 256 256,width=\expimgw]{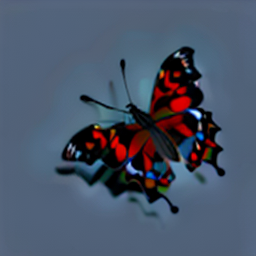}}
& {\includegraphics[bb=0 0 256 256,width=\expimgw]{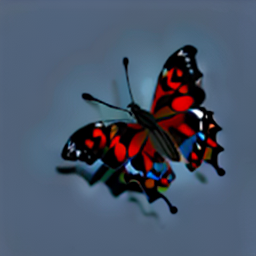}}
& {\includegraphics[bb=0 0 256 256,width=\expimgw]{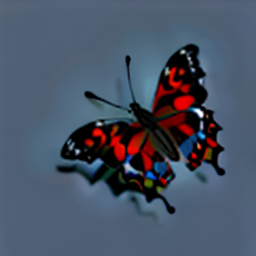}}
& {\includegraphics[bb=0 0 256 256,width=\expimgw]{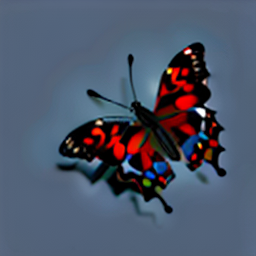}}

& {\includegraphics[bb=0 0 256 256,width=\expimgw]{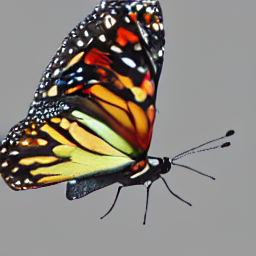}}
& {\includegraphics[bb=0 0 256 256,width=\expimgw]{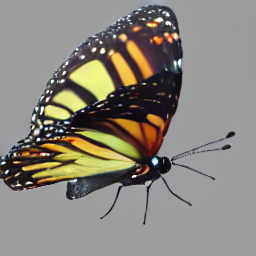}}
& {\includegraphics[bb=0 0 256 256,width=\expimgw]{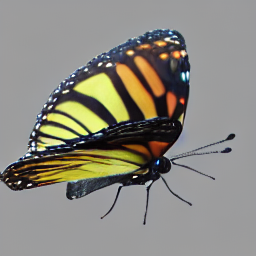}}
& {\includegraphics[bb=0 0 256 256,width=\expimgw]{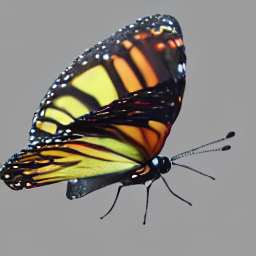}}
\\

\rotatebox{90}{\xspace\xspace View 1} 
& {\includegraphics[bb=0 0 256 256,width=\expimgw]{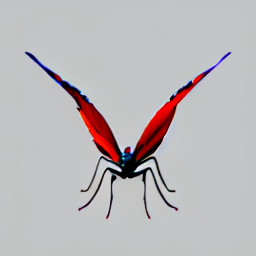}}

& {\includegraphics[bb=0 0 256 256,width=\expimgw]{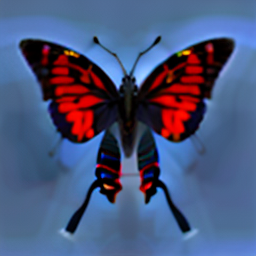}}
& {\includegraphics[bb=0 0 256 256,width=\expimgw]{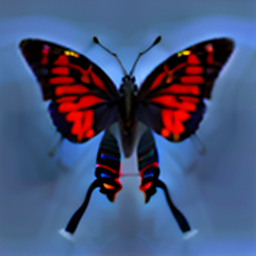}}
& {\includegraphics[bb=0 0 256 256,width=\expimgw]{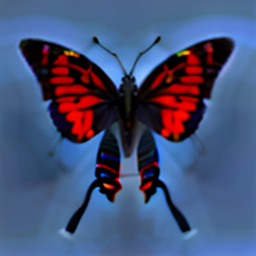}}
& {\includegraphics[bb=0 0 256 256,width=\expimgw]{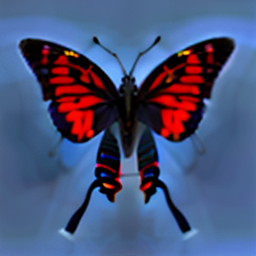}}

& {\includegraphics[bb=0 0 256 256,width=\expimgw]{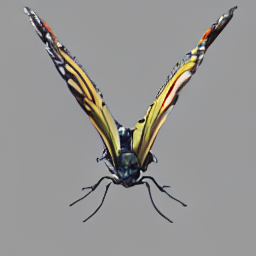}}
& {\includegraphics[bb=0 0 256 256,width=\expimgw]{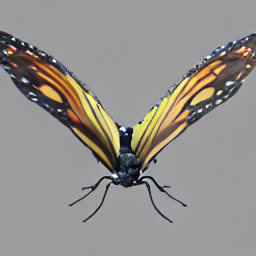}}
& {\includegraphics[bb=0 0 256 256,width=\expimgw]{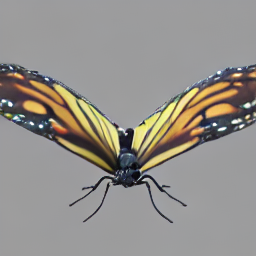}}
& {\includegraphics[bb=0 0 256 256,width=\expimgw]{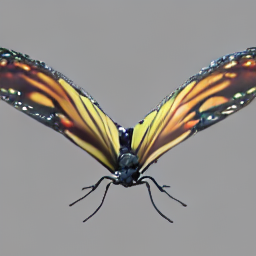}}
\\

\rotatebox{90}{\xspace\xspace View 2}
& {\includegraphics[bb=0 0 256 256,width=\expimgw]{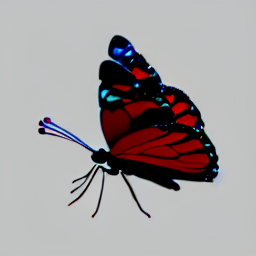}}

& {\includegraphics[bb=0 0 256 256,width=\expimgw]{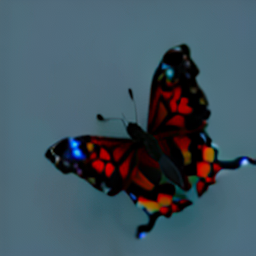}}
& {\includegraphics[bb=0 0 256 256,width=\expimgw]{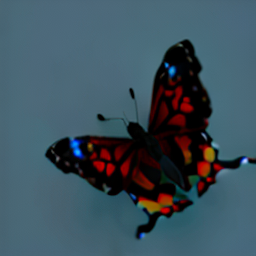}}
& {\includegraphics[bb=0 0 256 256,width=\expimgw]{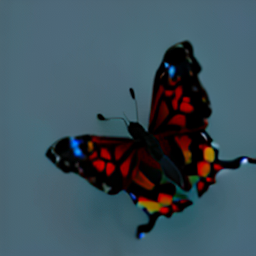}}
& {\includegraphics[bb=0 0 256 256,width=\expimgw]{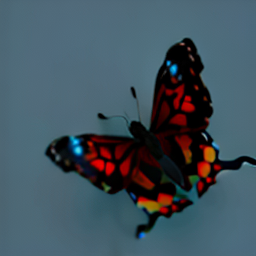}}

& {\includegraphics[bb=0 0 256 256,width=\expimgw]{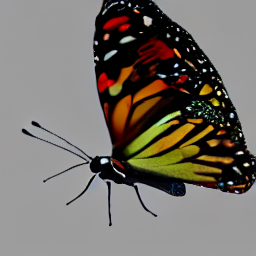}}
& {\includegraphics[bb=0 0 256 256,width=\expimgw]{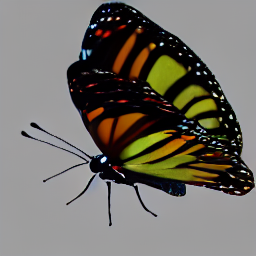}}
& {\includegraphics[bb=0 0 256 256,width=\expimgw]{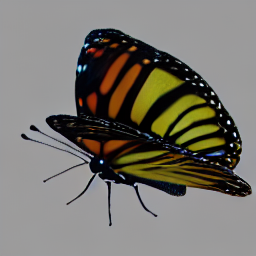}}
& {\includegraphics[bb=0 0 256 256,width=\expimgw]{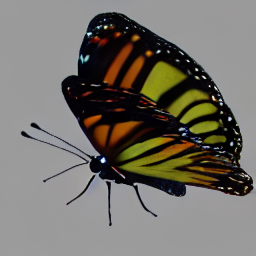}}
\\

\rotatebox{90}{\xspace\xspace View 3}
& {\includegraphics[bb=0 0 256 256,width=\expimgw]{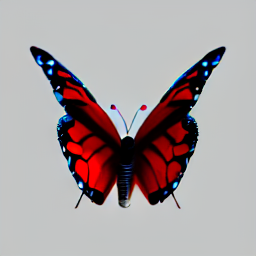}}

& {\includegraphics[bb=0 0 256 256,width=\expimgw]{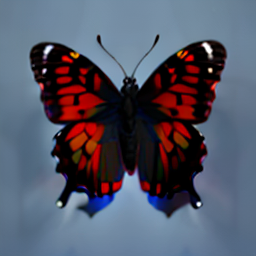}}
& {\includegraphics[bb=0 0 256 256,width=\expimgw]{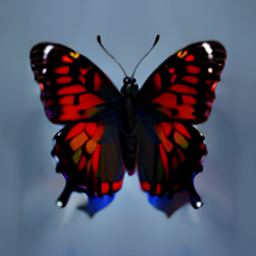}}
& {\includegraphics[bb=0 0 256 256,width=\expimgw]{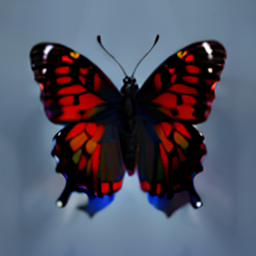}}
& {\includegraphics[bb=0 0 256 256,width=\expimgw]{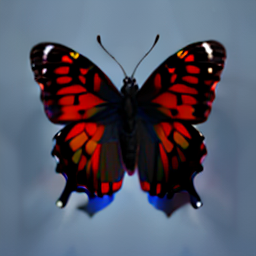}}

& {\includegraphics[bb=0 0 256 256,width=\expimgw]{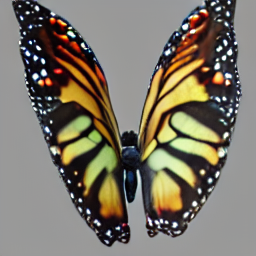}}
& {\includegraphics[bb=0 0 256 256,width=\expimgw]{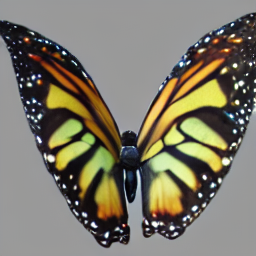}}
& {\includegraphics[bb=0 0 256 256,width=\expimgw]{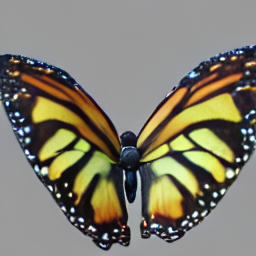}}
& {\includegraphics[bb=0 0 256 256,width=\expimgw]{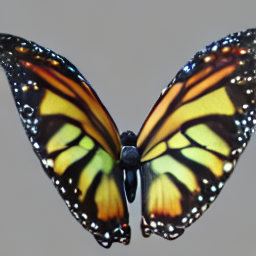}}
\\

\end{tabular}
\\
\textit{Text prompt: Beautiful, intricate butterfly, 3d asset.}
\caption{Comparison on \taskname generation. Although MVDream generates spatially 3D consistent images among views (the 1st column), MVDream + IP-AnimateDiff breaks the spatial 3D consistency among its generated videos. Instead, our method generates high-quality multi-view videos with large motions while maintaining temporal coherence and spatial 3D consistency.}
\vspace{-12pt}
\label{fig:com_butterfly}
\end{figure*}

\begin{figure*}[t]
\centering
\tabcolsep=0.02cm
\begin{tabular}{ccccc|cccc}
&\multicolumn{4}{c|}{w/o MS w SD} &\multicolumn{4}{c}{Ours}  \\
\rotatebox{90}{\xspace\xspace View 0}
& {\includegraphics[bb=0 0 256 256,width=0.11\textwidth]{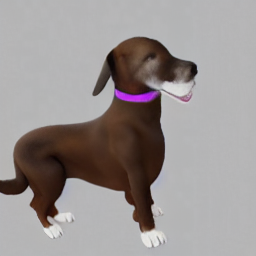}}
& {\includegraphics[bb=0 0 256 256,width=0.11\textwidth]{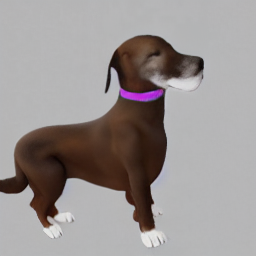}}
& {\includegraphics[bb=0 0 256 256,width=0.11\textwidth]{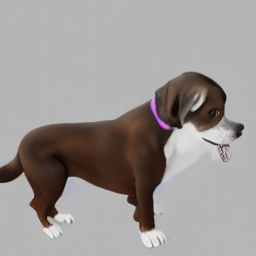}}
& {\includegraphics[bb=0 0 256 256,width=0.11\textwidth]{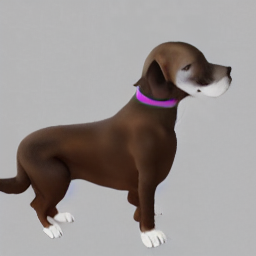}}

& {\includegraphics[bb=0 0 256 256,width=0.11\textwidth]{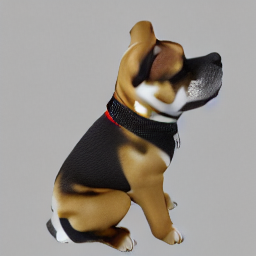}}
& {\includegraphics[bb=0 0 256 256,width=0.11\textwidth]{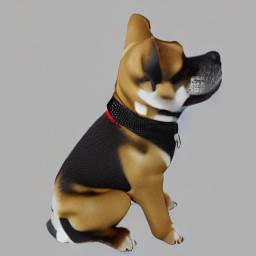}}
& {\includegraphics[bb=0 0 256 256,width=0.11\textwidth]{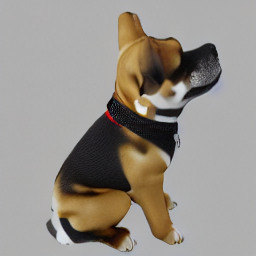}}
& {\includegraphics[bb=0 0 256 256,width=0.11\textwidth]{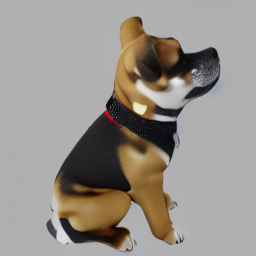}}
\\

\rotatebox{90}{\xspace\xspace View 1}
& {\includegraphics[bb=0 0 256 256,width=0.11\textwidth]{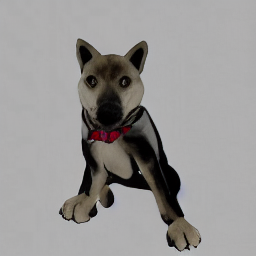}}
& {\includegraphics[bb=0 0 256 256,width=0.11\textwidth]{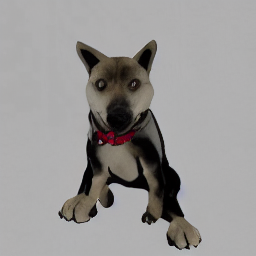}}
& {\includegraphics[bb=0 0 256 256,width=0.11\textwidth]{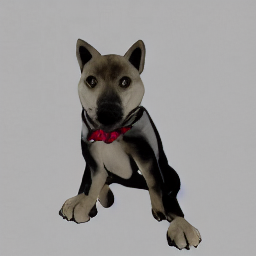}}
& {\includegraphics[bb=0 0 256 256,width=0.11\textwidth]{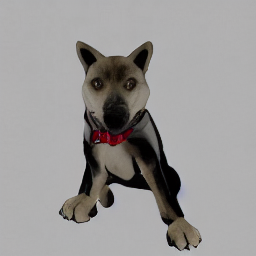}}

& {\includegraphics[bb=0 0 256 256,width=0.11\textwidth]{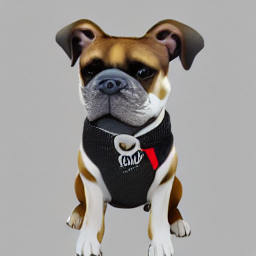}}
& {\includegraphics[bb=0 0 256 256,width=0.11\textwidth]{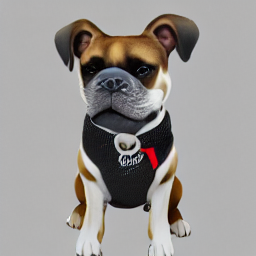}}
& {\includegraphics[bb=0 0 256 256,width=0.11\textwidth]{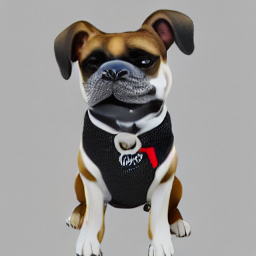}}
& {\includegraphics[bb=0 0 256 256,width=0.11\textwidth]{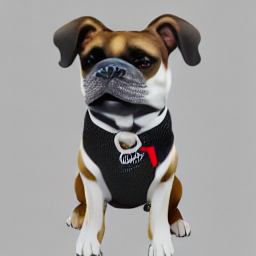}}
\\

\rotatebox{90}{\xspace\xspace View 2}
& {\includegraphics[bb=0 0 256 256,width=0.11\textwidth]{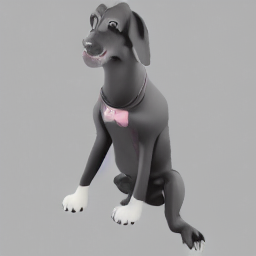}}
& {\includegraphics[bb=0 0 256 256,width=0.11\textwidth]{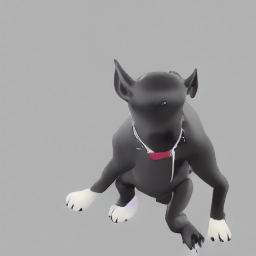}}
& {\includegraphics[bb=0 0 256 256,width=0.11\textwidth]{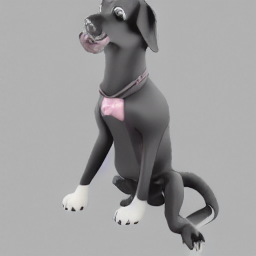}}
& {\includegraphics[bb=0 0 256 256,width=0.11\textwidth]{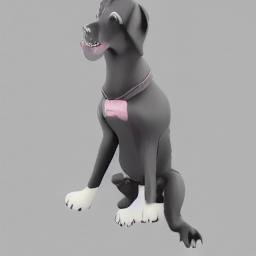}}

& {\includegraphics[bb=0 0 256 256,width=0.11\textwidth]{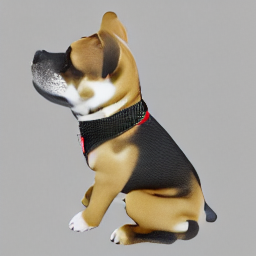}}
& {\includegraphics[bb=0 0 256 256,width=0.11\textwidth]{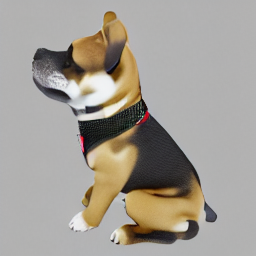}}
& {\includegraphics[bb=0 0 256 256,width=0.11\textwidth]{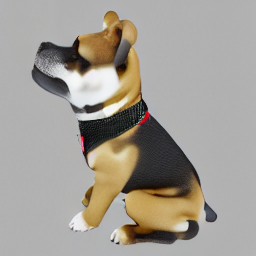}}
& {\includegraphics[bb=0 0 256 256,width=0.11\textwidth]{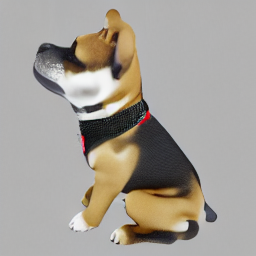}}
\\

\rotatebox{90}{\xspace\xspace View 3}
& {\includegraphics[bb=0 0 256 256,width=0.11\textwidth]{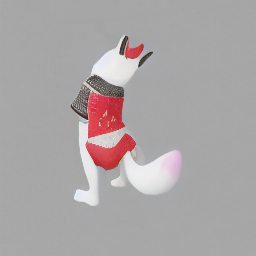}}
& {\includegraphics[bb=0 0 256 256,width=0.11\textwidth]{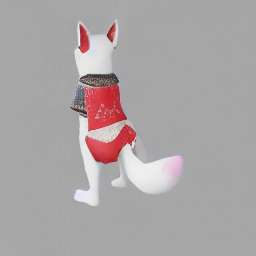}}
& {\includegraphics[bb=0 0 256 256,width=0.11\textwidth]{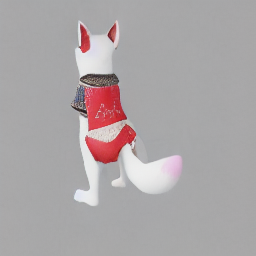}}
& {\includegraphics[bb=0 0 256 256,width=0.11\textwidth]{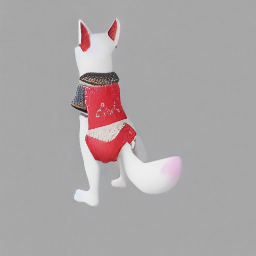}}

& {\includegraphics[bb=0 0 256 256,width=0.11\textwidth]{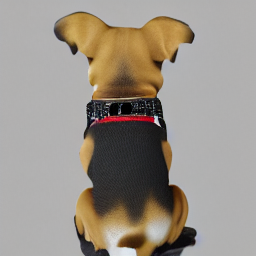}}
& {\includegraphics[bb=0 0 256 256,width=0.11\textwidth]{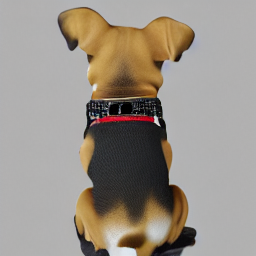}}
& {\includegraphics[bb=0 0 256 256,width=0.11\textwidth]{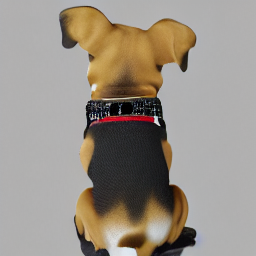}}
& {\includegraphics[bb=0 0 256 256,width=0.11\textwidth]{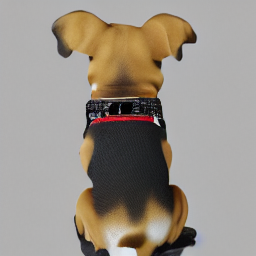}}
\\

\end{tabular}
\\

\textit{ Text prompt: a dog wearing an outfit, 3d asset}
\caption{Visual comparison of the contributions of our multi-view spatial module}
\label{fig:abla-spatial-3d}
\vspace{-13pt}
\end{figure*}

Given a text prompt, \textbf{MVDream + IP-AnimateDiff} generates multi-view videos in two stages, where MVDream generates multi-view images in the first stage, and IP-AnimateDiff animates each generated image from view into a 2D video in the second stage.   

Fig. \ref{fig:com_butterfly} and Tab. \ref{tab: compar} show that \textbf{MVDream + IP-AnimateDiff} achieves slightly better CLIP values. 
However,  our method outperforms \textbf{MVDream + IP-AnimateDiff} by a large margin in FVD and overall performance. 
 \textbf{MVDream + IP-AnimateDiff} introduces the noticeable 3D inconsistency among different views. For example, both appearances and motions of the butterfly in the view 0 video are inconsistent with those of view 3. 
 In contrast, our method not only achieves better performance in maintaining multi-view consistency, but also generates larger and more vivid motions for the butterfly, thanks to our pipeline and dataset. In addition, different from MVDream + IP-AnimateDiff employing two kinds of diffusion models and generating results in two stages, our method provides a unified diffusion model generating high-quality multi-view videos in only one stage.  Please refer to the Appendix for more results.

\begin{table*}[!t]
\centering
\caption{Multi-view video generation. Best in bold.}
\label{tab: compar}
\begin{tabular}{lccc}
\toprule  
Method & FVD  $\downarrow$  & CLIP $\uparrow$  &  Overall $\uparrow$  \\
\midrule 
 MVDream + IP-AnimateDiff& 2038.66 & \textbf{32.71} & 28\%\\
Ours & \textbf{1311.12} & 32.14 & \textbf{72\%}\\
\bottomrule 
\end{tabular}
\vspace{-15pt}
\end{table*}

\subsection{Ablation study and discussions}
We conduct the ablation study to show the effectiveness of the design in our multi-view spatial and temporal modules, as well as the proposed 3D-2D and 2D-3D alignment. 

\textbf{Design of multi-view spatial module.} We build a baseline named \textbf{w/o MS w SD} that employs original Stable Diffusion 1.5 \cite{sd15} as our multi-view spatial module and reuses its pre-trained weights.  We also insert the camera embedding into the Stable Diffusion to enable viewpoint control. That is, \textbf{w/o MS w SD} is to generate a single-view video (2D) conditioned on input text and camera poses. We train \textbf{w/o MS w SD} on our dataset, where single-view videos are used as training data.

Since single-view video generation is much simpler than multi-view video generation, \textbf{w/o MS w SD} achieves high performance in video quality. However, \textbf{w/o MS w SD}  fails to maintain multi-view consistency among different views (see Fig. \ref{fig:abla-spatial-3d}) and has degraded overall generation performance (see Tab. \ref{tab: abaltion}). For example, the motion and shapes of the dragon are significantly inconsistent among views.  Instead, by simply adapting a pre-trained multi-view image diffusion model as our spatial module, our method effectively ensures multi-view consistency.

\begin{figure*}
\centering
\tabcolsep=0.02cm
\begin{tabular}{ccccc|cccc}
&\multicolumn{4}{c|}{w/o MT w TM LoRA} & \multicolumn{4}{c}{Ours} \\
\rotatebox{90}{\xspace\xspace View 0}

& {\includegraphics[bb=0 0 256 256,width=0.11\textwidth]{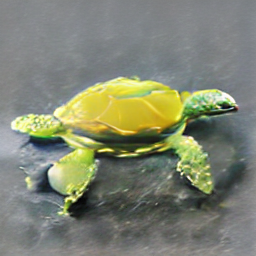}}
& {\includegraphics[bb=0 0 256 256,width=0.11\textwidth]{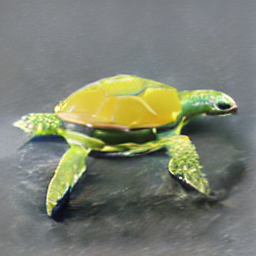}}
& {\includegraphics[bb=0 0 256 256,width=0.11\textwidth]{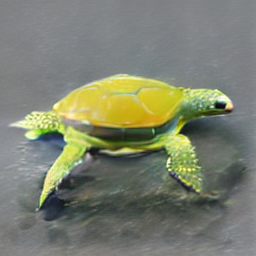}}
& {\includegraphics[bb=0 0 256 256,width=0.11\textwidth]{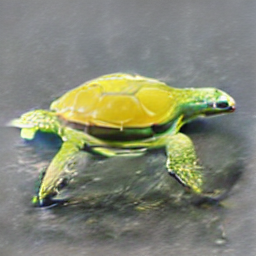}}

& {\includegraphics[bb=0 0 256 256,width=0.11\textwidth]{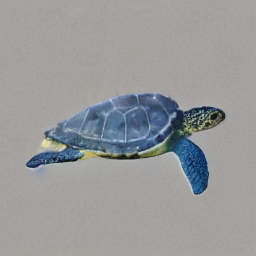}}
& {\includegraphics[bb=0 0 256 256,width=0.11\textwidth]{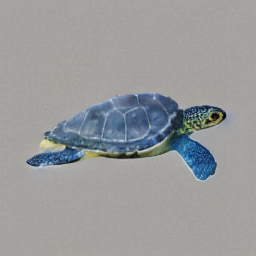}}
& {\includegraphics[bb=0 0 256 256,width=0.11\textwidth]{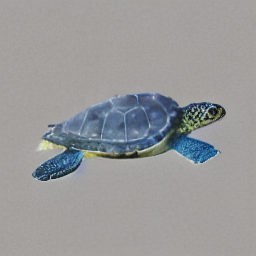}}
& {\includegraphics[bb=0 0 256 256,width=0.11\textwidth]{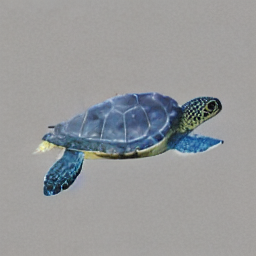}}

\\

\rotatebox{90}{\xspace\xspace View 1}

& {\includegraphics[bb=0 0 256 256,width=0.11\textwidth]{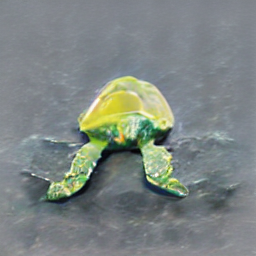}}
& {\includegraphics[bb=0 0 256 256,width=0.11\textwidth]{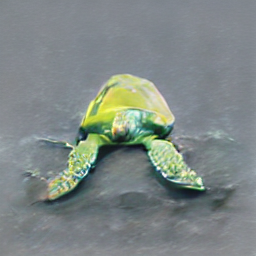}}
& {\includegraphics[bb=0 0 256 256,width=0.11\textwidth]{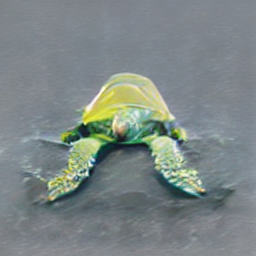}}
& {\includegraphics[bb=0 0 256 256,width=0.11\textwidth]{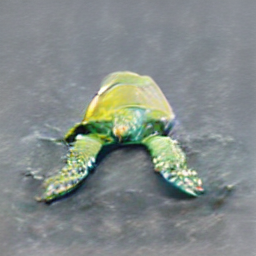}}

& {\includegraphics[bb=0 0 256 256,width=0.11\textwidth]{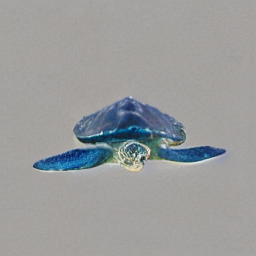}}
& {\includegraphics[bb=0 0 256 256,width=0.11\textwidth]{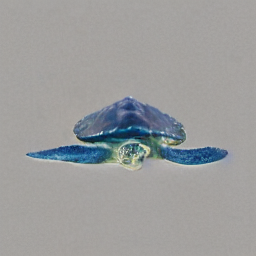}}
& {\includegraphics[bb=0 0 256 256,width=0.11\textwidth]{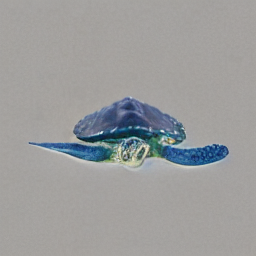}}
& {\includegraphics[bb=0 0 256 256,width=0.11\textwidth]{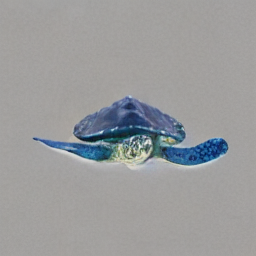}}

\\

\rotatebox{90}{\xspace\xspace View 2}
& {\includegraphics[bb=0 0 256 256,width=0.11\textwidth]{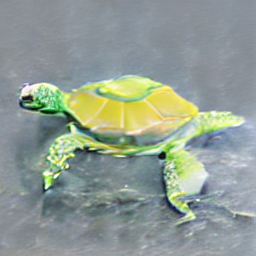}}
& {\includegraphics[bb=0 0 256 256,width=0.11\textwidth]{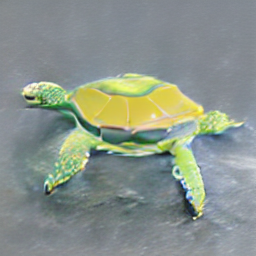}}
& {\includegraphics[bb=0 0 256 256,width=0.11\textwidth]{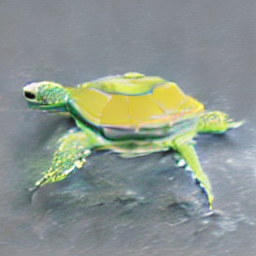}}
& {\includegraphics[bb=0 0 256 256,width=0.11\textwidth]{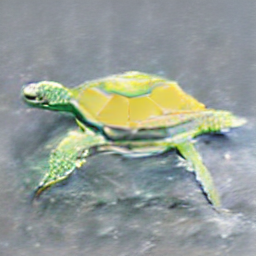}}

& {\includegraphics[bb=0 0 256 256,width=0.11\textwidth]{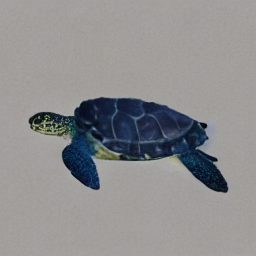}}
& {\includegraphics[bb=0 0 256 256,width=0.11\textwidth]{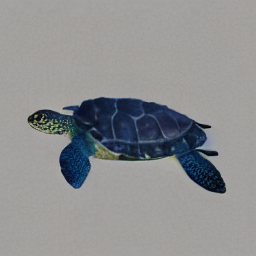}}
& {\includegraphics[bb=0 0 256 256,width=0.11\textwidth]{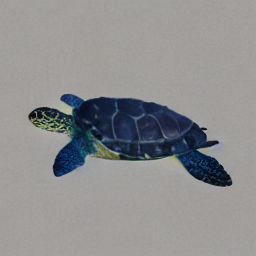}}
& {\includegraphics[bb=0 0 256 256,width=0.11\textwidth]{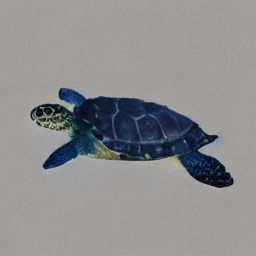}}

\\

\rotatebox{90}{\xspace\xspace View 3}
& {\includegraphics[bb=0 0 256 256,width=0.11\textwidth]{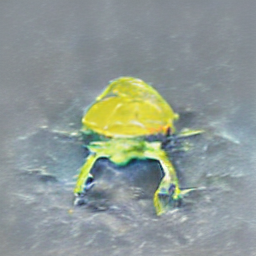}}
& {\includegraphics[bb=0 0 256 256,width=0.11\textwidth]{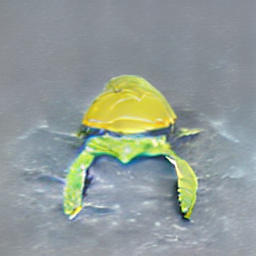}}
& {\includegraphics[bb=0 0 256 256,width=0.11\textwidth]{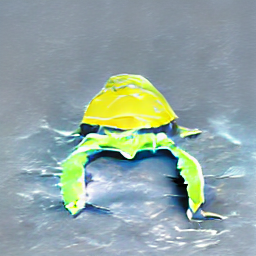}}
& {\includegraphics[bb=0 0 256 256,width=0.11\textwidth]{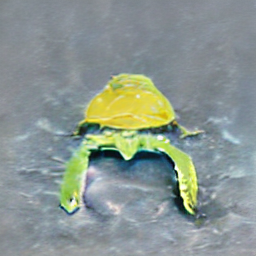}}

& {\includegraphics[bb=0 0 256 256,width=0.11\textwidth]{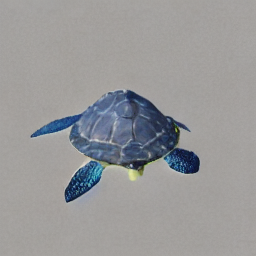}}
& {\includegraphics[bb=0 0 256 256,width=0.11\textwidth]{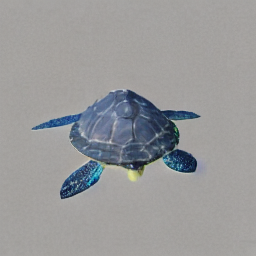}}
& {\includegraphics[bb=0 0 256 256,width=0.11\textwidth]{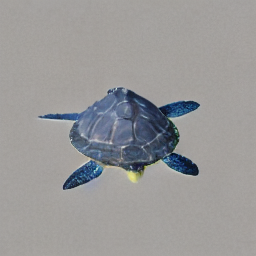}}
& {\includegraphics[bb=0 0 256 256,width=0.11\textwidth]{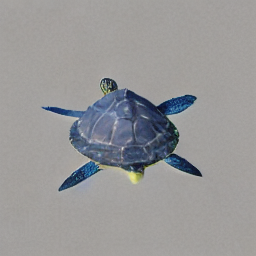}}
\end{tabular}
\\
\textit{ Text prompt: a sea turtle, 3d asset.}
\caption{Visual comparison of the contributions of our multi-view temporal  module}
\vspace{-20pt}
\label{fig:lora}
\end{figure*}
\textbf{Design of multi-view temporal module.} Recent methods apply LoRA \cite{lora} to the 2D temporal attention layers of a pre-trained 2D video diffusion model and fine-tune only LoRA for personalized and customized 2D video generation tasks \cite{animatediff,svd,ren2024customizeavideo}.
Following these methods, we build a temporal module named \textbf{TM LoRA} by inflating 2D temporal layers of AnimateDiff to handle multi-view videos and adding LoRA to the 2D temporal attention layers.
We replace our multi-view temporal module with \textbf{TM LoRA}, and denote it by \textbf{w/o MT w TM LoRA}.
Fig. \ref{fig:lora} and Tab. \ref{tab: abaltion} shows  \textbf{w/o MT w TM LoRA} generates low-quality results, despite being fine-tuned on our dataset.
Instead, our multi-view temporal module inserts 3D-2D alignment and 
 2D-3D alignment layers before and after the 2D temporal attention layers, enabling the multi-view temporal module to be compatible with the multi-view spatial module.

\begin{wraptable}[9]{r}{0.40\textwidth}
\vspace{-21pt}
\caption{The ablation study results. The overall performance is assessed by a user study using paired comparison \cite{4dfy,david1963method}.}%
\resizebox{0.35\textwidth}{!}{ %
\begin{tabular}{lccc}
\toprule  
Method &   Overall $\uparrow$    \\
\midrule
w/o MS w SD &  44.25\% \\ 
w/o MT w TM LoRA &  10.00\% \\
w/o 3D-2D alignment &  53.50\% \\
w/o 2D-3D alignment &   55.50\%\\
Ours &    \textbf{81.00\%}\\
\bottomrule 
\end{tabular}
} %
\label{tab: abaltion}
\end{wraptable}

\textbf{Effect of 3D-2D  alignment.}  We remove the proposed 3D-2D alignment from our model and train the model on our dataset with the same settings.  Tab. \ref{tab: abaltion} shows \textbf{w/o 3D-2D alignment} degrades our temporal coherence and video quality performance. Instead, by projecting the feature to the latent space of the pre-trained 2D attention layers, our 3D-2D alignment layer effectively enables the 2D attention layers to align temporally correlated content, ensuring the video quality and temporal coherence. %

\textbf{Effect of 2D-3D  alignment.}  As shown in Tab. \ref{tab: abaltion}, ``w/o 2D-3D temporal alignment" achieves lower performance with the same training settings due to the removal of 2D-3D temporal alignment. The results indicate that only 3D-2D  alignment is insufficient in jointly leveraging the pre-trained 2D temporal layers \cite{animatediff} and the multi-view image diffusion model \cite{MVDream} in our diffusion model. Instead, our 2D-3D  alignment projects the features processed by the pre-trained 2D temporal layers back to the latent space of the multi-view image diffusion model, leading to high-quality results.

\textbf{Training cost.} \label{sec:train_cost} 
MVDream is trained on 32 Nvidia Tesla A100 GPUs, which takes 3 days, and AnimateDiff takes around 5 days on 8 A100 GPUs. By combining and reusing the layers of MVDream and AnimateDiff, our method only needs to train the proposed 3D-2D alignment and 2D-3D layers, reducing the training cost to around 2 days with 8 A100 GPUs.

\section{Conclusions} \label{sec:conclusion}

In this paper, we propose a novel diffusion-based pipeline named Vivid-ZOO that generates high-quality multi-view videos centered around a dynamic 3D object from text. The presented multi-view spatial module ensures the multi-view consistency of generated multi-view videos, while the multi-view temporal module effectively enforces temporal coherence.  By introducing the proposed 3D-2D temporal alignment and 2D-3D temporal alignment layers, our pipeline effectively leverages the layers of the pre-trained multi-view image and 2D video diffusion models, reducing the training cost and accelerating the training of our diffusion model.  We also construct a dataset of captioned multi-view videos, which facilitates future research in this emerging area.

\clearpage
\bibliographystyle{plain}
\bibliography{refer}
\clearpage

\end{document}